\documentclass{article}
\usepackage{ijcai23}

\usepackage{times}
\usepackage{soul}
\usepackage{url}
\usepackage[hidelinks]{hyperref}
\usepackage[utf8]{inputenc}
\usepackage[small]{caption}
\usepackage{graphicx}
\usepackage{amsmath}
\usepackage{amsthm}
\usepackage{amssymb}
\usepackage{booktabs}
\usepackage{algorithm}
\usepackage{algorithmic}
\usepackage[switch]{lineno}
\usepackage{dsfont}
\usepackage{pifont}
\usepackage{color}

\title{Complete Instances Mining for Weakly Supervised Instance Segmentation}

\author{
Zecheng Li$^1$\and
Zening Zeng$^1$\and
Yuqi Liang$^1$\And
Jin-Gang Yu$^{1,2}$\thanks{Corresponding Author}
\affiliations
$^1$South China University of Technology\\
$^2$Pazhou Laboratory\\
\emails
lizecheng19@gmail.com,
\{zeningzeng, auyqliang\}@mail.scut.edu.cn,
jingangyu@scut.edu.cn
}

\begin{document}

\maketitle
\renewcommand{\thefootnote}{}
\footnotetext{The copyrighted version of the paper is at \href{https://www.ijcai.org/proceedings/2023/127}{ijcai.org/proceedings/2023/127}.}

\begin{abstract}
Weakly supervised instance segmentation (WSIS) using only image-level labels is a challenging task due to the difficulty of aligning coarse annotations with the finer task. However, with the advancement of deep neural networks (DNNs), WSIS has garnered significant attention. Following a proposal-based paradigm, we encounter a redundant segmentation problem resulting from a single instance being represented by multiple proposals. For example, we feed a picture of a dog and proposals into the network and expect to output only one proposal containing a dog, but the network outputs multiple proposals. To address this problem, we propose a novel approach for WSIS that focuses on the online refinement of complete instances through the use of MaskIoU heads to predict the integrity scores of proposals and a Complete Instances Mining (CIM) strategy to explicitly model the redundant segmentation problem and generate refined pseudo labels. Our approach allows the network to become aware of multiple instances and complete instances, and we further improve its robustness through the incorporation of an Anti-noise strategy. Empirical evaluations on the PASCAL VOC 2012 and MS COCO datasets demonstrate that our method achieves state-of-the-art performance with a notable margin. Our implementation will be made available at \href{https://github.com/ZechengLi19/CIM}{https://github.com/ZechengLi19/CIM}.
\end{abstract}

\section{Introduction}
Instance segmentation involves the simultaneous estimation of object location and masking segmentation, and it has made significant progress with the assistance of large datasets and instance-level annotations. However, process of obtaining instance-level annotations can be costly and time-consuming. As a solution, weak annotations such as box-level and image-level annotations have been utilized in instance segmentation. Among these weak annotations, the use of image-level annotations is the most cost-effective, but also the most challenging due to the difficulty of aligning coarse annotations with the finer task.

\begin{figure}[t]
    \begin{center}
    \includegraphics[width=1\linewidth]{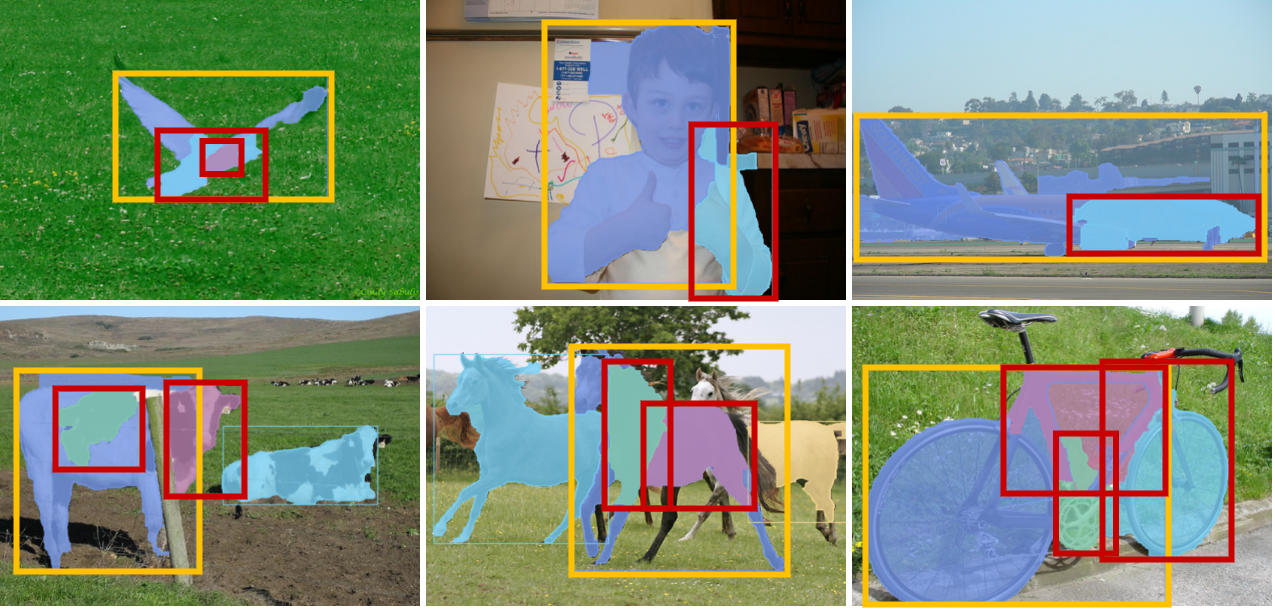}
    \end{center}
    \caption{Redundant segmentation. For each instance, it always corresponds to multiple proposals. \textcolor[RGB]{255,193,0}{Yellow boxes:} expected segmentations. \textcolor[RGB]{200,0,0}{Red boxes:} redundant segmentations.}
    \label{fig:motivation}
\end{figure}

Weakly supervised instance segmentation (WSIS), which involves using only image-level annotations, has experienced widespread growth in recent years. There are two main paradigms for addressing WSIS: proposal-based paradigm and proposal-free paradigm. The proposal-based paradigm~\cite{zhou2018weakly,liu2020leveraging,shen2021parallel,ou2021ws} assumes that an instance can be represented by a proposal, thus simplifying WSIS as a classification task. However, these approaches will result in impaired segmentation performance due to redundant segmentation
, as illustrated in Figure~\ref{fig:motivation}. Typically, if we simply select a few results as output, the segmentation results will be the most discriminative parts of instances. In contrast, the proposal-free paradigm~\cite{ahn2019weakly,kim2022beyond} generates segmentation results online without proposals. During training, this paradigm relies on confident pre-computed pseudo labels to achieve better performance. To leverage these pre-computed pseudo labels, IRNet~\cite{ahn2019weakly} utilizes Class Attention Maps (CAM)~\cite{zhou2016CAM} while BESTIE~\cite{kim2022beyond} uses weakly supervised semantic segmentation (WSSS) maps. The proposal-free paradigm is heavily dependent on pre-computed pseudo labels, which can limit its potential for further improvement.

To address the limitations mentioned above, we propose a novel proposal-based method that operates in an online refinement manner and consists of three key components: MaskIoU heads, a Complete Instances Mining (CIM) strategy, and an Anti-noise strategy. Our method also utilizes pre-computed pseudo labels to warm up the model. Instead of using techniques such as random walk (RW)~\cite{ahn2019weakly} or conditional random fields (CRF)~\cite{ge2019label,shen2019cyclic,zhu2019learning} to mine complete instances, which can significantly slow down training process, our method employs MaskIoU heads to predict the integrity scores of proposals and the CIM strategy to mine complete instances. We then generate refined pseudo labels based on spatial relationships using the mined complete instances. Finally, the Anti-noise strategy ensures that our method does not suffer from significant performance degradation when pre-computed/refined pseudo labels are noisy.

The main contributions of our work are as follows: 
\begin{itemize}
    \item 
    We introduce the MaskIoU head to WSIS for the first time and propose an Anti-noise strategy to filter out noise caused by pre-computed pseudo labels and refined pseudo labels, improving the robustness of our method.
    \item 
    We explicitly address the problem called redundant segmentation and present an effective Complete Instances Mining (CIM) strategy to guide the network to pay more attention to complete instances.
    \item 
    Despite its simplicity, our method achieves state-of-the-art performance on the PASCAL VOC 2012 and MS COCO datasets with a notable margin.
\end{itemize}

\section{Related Work}
\subsection{Weakly Supervised Object Detection}
Multiple Instance Learning (MIL) is a popular paradigm for addressing weakly supervised object detection (WSOD), as it utilizes proposals generated by selective search~\cite{uijlings2013selective} as a bag of instances.

WSDDN~\cite{bilen2016weakly} proposed a classification branch and a detection branch to output confident proposals as results, but this approach tends to predict the most discriminative parts of objects. To handle this issue,  contextlocnet~\cite{kantorov2016contextlocnet} introduced contextual information to a contrastive model. OICR~\cite{Tang_2017_CVPR} proposed an online instance classifier refinement strategy, which utilizes the proposals of the highest confidence as a pseudo label to supervise the next branch for improved performance. MIST~\cite{ren-cvpr2020} considered the presence of multiple instances and employed Concrete DropBlock to mine complete instances. C-MIL~\cite{wan2019c} designed a novel MIL loss to avoid getting stuck into local minima, while PCL~\cite{Tang2020pcl} transformed the single MIL problem into multiple MIL subproblems through clustering proposals.  WSOD$\mathrm{}^2$~\cite{zeng2019wsod2} combined bottom-up and top-down objectness knowledge as evidence to discover complete instances in the candidates. NDI-WSOD~\cite{wang2022absolute} and OD-WSCL~\cite{seo2022Contrastive} approached WSOD through contrastive learning.

Although WSIS is distinct from WSOD, it is worth noting that our approach was inspired by WSOD approaches and WSOD approaches can be easily transferred to WSIS.

\subsection{Weakly Supervised Instance Segmentation}
WSIS with only image-level annotations can be divided into two paradigms: proposal-based paradigm and proposal-free paradigm.

Following the proposal-based paradigm, PRM~\cite{zhou2018weakly} used peaks in Peak Response Maps (PRM) as instance cues and combined them with CAM to predict instances. IAM~\cite{zhu2019learning} generated Instance Activation Maps (IAM) through an extent filling module to identify complete instances. Label-PEnet~\cite{ge2019label} employed a cascaded pipeline in a coarse-to-fine manner to simultaneously perform classification, object detection, and instance segmentation. Arun et al.~\cite{arun2020weakly} used a conditional distribution to explicitly model the uncertainty, which enables pseudo labels to become more reliable. Fan et al.~\cite{fan2018associating} and LIID~\cite{liu2020leveraging} used instance saliency labels to support the generation of pseudo labels and considered the relationships in the training set. WS-RCNN~\cite{ou2021ws} proposed an Attention-Guided Pseudo Labeling (AGPL) strategy and an Entropic OpenSet Loss to improve WSIS. PDSL~\cite{shen2021parallel} treated proposals as segmentation cues and proposed a framework for learning detection and segmentation in parallel. Our proposed method is similar to PDSL, but does not require additional segmentation branches due to the ability of proposals to effectively segment instances.

Following the proposal-free paradigm, IRNet~\cite{ahn2019weakly} used displacement field to indicate instances and generate complete instances via Class Boundary Maps and random walk. BESTIE~\cite{kim2022beyond} transferred the knowledge of WSSS to WSIS by a strengthening constraint. While our method also leverages pre-computed pseudo labels, we design our approach in an online refinement manner and propose an Anti-noise strategy to reduce the dependence on these pre-computed labels.

Thorough comparisons of representative proposal-based and proposal-free methods show that outputs from these two paradigms focus on different aspects. Proposal-based methods tend to have higher recall and lower precision, while proposal-free methods exhibit the opposite. Despite these differences, there is not a significant gap in performance between the two paradigms. Previous proposal-based methods have effectively utilized the power of proposals, which not only consider bottom-up information of the image but also simplify the task. However, a convolutional neural network (CNN) trained for image classification typically results in redundant segmentation. To alleviate this problem, some researchers~\cite{zhu2019learning,arun2020weakly,ou2021ws} have made some attempts. In contrast, our method explicitly models this problem and addresses it through the Complete Instances Mining (CIM) strategy.

\begin{figure*}[t]
\begin{center}
\includegraphics[width=0.95\linewidth]{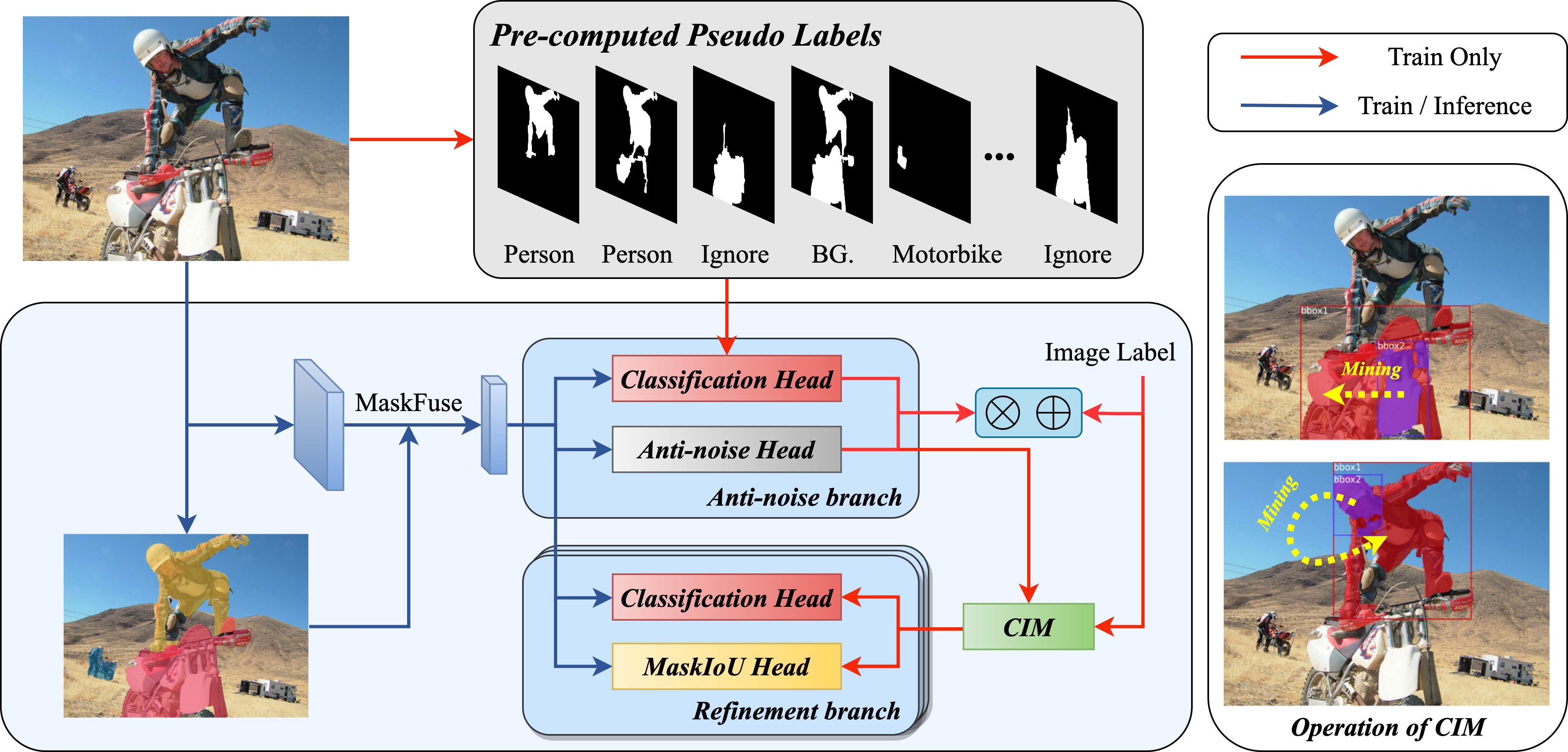}
\end{center}
   \caption{Overview of our proposed method. Our framework mainly contains three components: an Anti-noise branch, $K$ Refinement branches, and Complete Instances Mining (CIM) strategy. Proposal features are generated by MaskFuse and forked into multiple branches. Both Anti-noise and Refinement branches output classification and integrity scores. CIM leverages output of preceding branch to generate refined pseudo labels to supervise next branch, while Anti-noise branch is supervised by pre-computed pseudo labels.
   In the right column, purple and red represent seeds and pseudo ground truth, respectively. The seeds spread in space to find complete proposals as pseudo ground truth through spatial relationships and integrity scores.}
\label{fig:overview}
\end{figure*}

\section{Proposed Method}
\subsection{Preliminary and Overview}
Given an image \textit{\textbf{I}} sized by $H_{I} \times W_{I}$, and its image-level label $Y \in \mathbb{R}^{C}$, where $C$ is the number of categories and $Y_c=1$ indicates that image \textit{\textbf{I}} contains the $c^{th}$ category, and otherwise, $Y_c=0$. To simplify the task, proposals $R \in \mathbb{R}^{N \times H_{I} \times W_{I}}$ are obtained using off-the-shelf proposal techniques~\cite{pont2016multiscale,maninis2017convolutional}, where $R_n$ is a binary segmentation mask and $N$ indicates the number of proposals. In this work, we modify the image-level label $Y \in \mathbb{R}^{C+1}$ to include the background category, denoted as $Y_0=1$. 

Our method follows the proposal-based paradigm and consists of an Anti-noise branch and several Refinement branches, as illustrated in Figure~\ref{fig:overview}. 
To begin, we obtain pre-computed pseudo labels using the AGPL strategy proposed by WS-RCNN~\cite{ou2021ws}. Then, we use MaskFuse to extract proposal features and feed them into multiple branches. Finally, the CIM strategy explicitly models the redundant segmentation problem and generates refined pseudo labels. To further enhance the robustness of our network, we apply an Anti-noise strategy.

\subsection{Reviewing AGPL}
\label{section:Overview of AGPL}
The AGPL strategy is a method for generating pseudo labels for a set of proposals in the WS-RCNN model. The method starts by training a classifier and producing CAM. For each target category $c$, confident peaks from CAM are selected as instance cues, denoted as $p^c=\{p^c_1, p^c_2, \cdots ,p^c_i\}$, where $p^c_i$ means the $i^{th}$ peak in the $c^{th}$ category. For each peak, AGPL averages and thresholds proposals containing it to leverage support mask $S^c_i$, denoted the number of proposals as $n_i^c$.
\begin{equation}
    S^c_i = \left ( \frac{1}{n^c_i} \sum_{p^c_i \in R_n} R_n \right) > 0.7,
\end{equation}
The AGPL strategy sorts the support mask set in descending order according to scores of peaks and assigns labels to proposals based on the IoU between the proposal and the support mask. The pre-computed pseudo labels of proposals are represented as $\hat{y}^{0} \in \mathbb{R}^{N \times \{C+1\}}$.
\begin{equation}
    \begin{array}{cc}
    \hat{y}^{0}_{i,c} = 1 & \text{if} \ \text{IoU} (R_i,S^c_k)>0.5 
    \end{array}
\end{equation}
During the labeling process, each proposal is assigned to only one category. Proposals that overlap with support masks but are not assigned to any categories are assigned as background. Finally, proposal clusters are generated based on whether proposals are assigned by the same support mask, with each background proposal being assigned as a single cluster. The proposal clusters set is denoted as $H=\{ (\mathcal{C}_1, M_1, s_1), (\mathcal{C}_2, M_2, s_2),\cdots, (\mathcal{C}_{N^0}, M_{N^0}, s_{N^0})\}$, where $\mathcal{C}_n$, $M_n$, and $s_{n}$ represent the set of proposals, the number of proposals, and the label of the $n^{th}$ proposal cluster, respectively.

\subsection{MaskFuse}
In contrast to WSOD, WSIS utilizes binary masks as proposals rather than bounding boxes. As a result, RoIPool~\cite{girshick2015fast} and RoIAlign~\cite{he2017mask} operations are not available for WSIS. 

To address the lack of feature extraction operation in WSIS, we propose a lightweight operation called MaskFuse. The MaskFuse leverages the bounding box feature $B \in \mathbb{R}^{h \times w \times D}$ obtained through the RoIAlign operation and extracts the corresponding proposal ${R_{crop}} \in \mathbb{R}^{h \times w}$ using the RoICrop operation. Then, we concatenate ${R_{crop}} \odot B $ with $B$ and utilize a convolutional layer and two fully connected layers to fuse features from mask-level and box-level. The MaskFuse allows each proposal to be represented by a feature with contextual information, enabling WSOD methods to be transferred to WSIS.

\subsection{Refinement Branch}

The MaskIoU head, proposed by MS R-CNN~\cite{huang2019mask}, and the center-ness head, proposed by FCOS~\cite{tian2019fcos}, are designed to learn the quality of predicted results (i.e., masks and bounding boxes) in their respective methods. However, this powerful head is missing in WSIS. To fill this gap, we implement the MaskIoU head into our framework. 


Given proposal features, the $k^{th}$ Refinement branch produces two matrixes $y^{k},t^{k} \in \mathbb{R}^{N \times \{C+1\}}$ from classification and MaskIoU heads, which represent the classification and integrity scores, respectively. Furthermore, we employ the CIM strategy to produce refined pseudo labels from preceding branch, denoted as $\hat{y}^{k}, \hat{t}^{k}$. 
Following OICR~\cite{Tang_2017_CVPR}, we also propose loss weights $w^k \in \mathbb{R}^{N}$ to mitigate the degradation caused by noisy refined pseudo labels.
\begin{align}
    \mathcal{L}_{ref}^{k} = &-\frac{1}{N_{fg} + N_{bg}} \sum^{N}_{n=1} \sum^{C}_{c=0} w^k_n \ \hat{y}^{k}_{n,c} \log y^{k}_{n,c} \nonumber \\
    &+ \frac{1}{N_{fg}}  \sum^{N}_{n=1} \sum^{C}_{c=1}  w^k_n\ \hat{y}^{k}_{n,c} \mathcal{L}_{Smooth-L1} (\hat{t}^{k}_{n,c} - t^{k}_{n,c}) \nonumber \\
\end{align}
where $N_{fg}$ and $N_{bg}$ represent number of foreground and background. $\mathcal{L}_{Smooth-L1}$ indicates smooth-L1 loss~\cite{girshick2015fast}.

\begin{algorithm}[t]
    \caption{Complete Instances Mining (CIM) strategy}
    \label{alg:CIM}
    \textbf{Input}: Image label $Y$, proposals $R$, classification scores $y^{k-1}$, integrity scores $t^{k-1}$ \\
    \textbf{Parameter}: NMS threshold $\tau_{nms}$, containment threshold $\tau_{con}$, percentage of seeds $p_{seed}$  \\
    \textbf{Output}: Pseudo ground truth $P^k$
    \begin{algorithmic}[1] 
        \STATE $P^k = \varnothing$
        \FOR{$c=1$ to $C$}
            \IF{$Y_c=1$}
                \STATE // \textbf{Step 1: selecting seeds}
                \STATE $R_{cls} \leftarrow Sort (y^{k-1}_{*,c})$ 
                \STATE $R_{keep} \leftarrow $ keep top $p_{seed}$ percent of  $R_{cls}$
                \STATE $ind_{seed}=NMS (R_{keep}, \tau_{nms})$
                \STATE // \textbf{Step 2: mining pseudo ground truth}
                \STATE Calculate matrix $M \in \mathbb{R}^{N \times N}, M_{i,j} = \frac{R_{i} \cap R_{j}}{R_{j}}$
                \STATE $M_{con}=M[:,ind_{seed}] > \tau_{con}$ 
                \STATE $ind_{gt} = argmax (M_{con} t^{k-1}_{*,c},dim=0)$
                \STATE $P^{k}_{c}.append(R[ind_{gt}])$
            \ENDIF
        \ENDFOR
    \end{algorithmic}
\end{algorithm}
\subsection{Complete Instances Mining (CIM)}


Motivated by redundant segmentation, we propose a novel Complete Instances Mining (CIM) strategy. CIM can be divided into two steps: selecting seeds and mining pseudo ground truth. The first step is similar to the operation of MIST~\cite{ren-cvpr2020} strategy.

Let's start with a simple modeling of redundant segmentation. As we expected, partial segmentations typically have high classification scores and low integrity scores, while complete segmentations have the opposite. Additionally, partial segmentations are often spatially contained within complete segmentations. Hence, we utilize classification and integrity scores as metrics for CIM, illustrated in Algorithm~\ref{alg:CIM}. CIM awares multiple instances through selecting seeds, which maintains a high recall of pseudo ground truth. For each seed, the proposal that contains it and has the highest integrity score is considered as its corresponding pseudo ground truth. Mining pseudo ground truth enables our method to identify complete instances by aggregating multiple seeds.

To collect more pseudo labels, we assign refined pseudo labels to all proposals. For each proposal, if it highly overlaps with pseudo ground truth, we consider its classification target to be the category of pseudo ground truth. $\tau_{cls}$ is a hyper-parameter.
\begin{equation}
\label{eq:hat_y_cls}
    \hat{y}^{k}_{i,c} = \left \{ 
    \begin{array}{cc}
      1  &  \text{if} \ \max (\text{IoU} (R_i, P^{k}_{c})) > \tau_{cls} \\
      0  &  \text{otherwise}
    \end{array}
    \right.
\end{equation}
The integrity target is determined in the same way. $\tau_{iou}$ is also a hyper-parameter.
\begin{equation}
\label{eq:hat_y_iou}
\hat{t}^{k}_{i,c} = \left \{ 
    \begin{array}{cc}
      1  &  \text{if} \ \max (\text{IoU} (R_i, P^{k}_{c})) > \tau_{iou} \\
      0  &  \text{otherwise}
    \end{array}
    \right.
\end{equation}
In which, $\tau_{iou}$ is typically greater than $\tau_{cls}$. The loss weight for each proposal can be calculated using Equation~\ref{eq:w^k}, where $c$ and $j$ indicate the category and index of pseudo ground truth that has the largest overlap with the $R_{i}$, respectively.
\begin{equation}
\label{eq:w^k}
    w^{k}_i = y^{k-1}_{j,c} \ t^{k-1}_{j,c}
\end{equation}
We follow the rule mentioned in Section~\ref{section:Overview of AGPL} to assign background labels. 
To further improve performance, we implement a cascaded threshold $\tau_{cas}$ into our framework inspired by Cascade R-CNN~\cite{Cai2018Cascade}. This involves modifying $\tau_{cls}$ and $\tau_{iou}$ in the $k^{th}$ Refinement branch to $\tau_{cls}+ (k-1)\ \tau_{cas}$ and $\tau_{iou}+ (k-1)\ \tau_{cas}$, respectively.

\begin{table*}[t!]
    \centering
    \begin{tabular}{c|c|c|c c c c}
        \toprule
        Method & Backbone & Sup. & $mAP_{25}$ & $mAP_{50}$ & $mAP_{70}$ & $mAP_{75}$ \\
        \midrule
        Mask R-CNN~\cite{he2017mask}  & ResNet-101         & $\mathcal{M}$ & 
        76.7 & 67.9 & 52.5 & 44.9 
        \\
        
        \midrule
        PRM~\cite{zhou2018weakly}     & ResNet-50          & $\mathcal{I}$  & 
        44.3 & 26.8 & - & 9.0
        \\
        IAM~\cite{zhu2019learning}    & ResNet-50          & $\mathcal{I}$  &
        45.9 & 28.3 & - & 11.9
        \\

        MIST$^*$~\cite{ren-cvpr2020}   & VGG-16       
        & $\mathcal{I}$  & 
        58.5 & 43.1 & 23.2 & 18.3
        \\
        
        Label-PEnet~\cite{ge2019label} & VGG-16            & $\mathcal{I}$   & 
        49.2 & 30.2 & - & 12.9
        \\
        
        WS-RCNN~\cite{ou2021ws} & VGG-16 & $\mathcal{I}$ &
        57.2 & 42.7 & - & 19.4
        \\
        
        PDSL~\cite{shen2021parallel}  & ResNet50-WS        & $\mathcal{I}$    &
        59.3 & 49.6 & - & 12.7
        \\
        
        BESTIE~\cite{kim2022beyond}   & HRNet-W48           & $\mathcal{I}$  & 
        53.5 & 41.8 & 28.3 & 24.2
        \\

        \midrule
        Ours             & ResNet-50  
        & $\mathcal{I}$  & 
        64.9 & 51.1 & 32.4 & 26.1
        \\
        Ours             & VGG-16 
        & $\mathcal{I}$  & 
        65.6 & 50.8 & 31.0 & 25.2
        
        \\
        Ours             & HRNet-W48 
        & $\mathcal{I}$ &
        \textbf{68.3} & \textbf{52.6} & \textbf{33.7} & \textbf{28.4}

        \\
        \midrule
        
        WISE$^{\dagger}$~\cite{laradji2019masks}  & ResNet-50    & $\mathcal{I}$  &
        49.2 & 41.7 & - & 23.7
        \\
        IRNet$^{\dagger}$~\cite{ahn2019weakly}    & ResNet-50  
        & $\mathcal{I}$  & 
        - & 46.7 & 23.5 & -
        \\
        LIID$^{\dagger}$~\cite{liu2020leveraging} & ResNet-50    & $\mathcal{I,S}$ & 
        - & 48.4 & - & 24.9
        \\
        Arun et al.$^{\dagger}$~\cite{arun2020weakly} & ResNet-50      & $\mathcal{I}$      &
        59.7 & 50.9 & 30.2 & 28.5
        \\ 
        WS-RCNN$^{\dagger}$~\cite{ou2021ws} & VGG-16 & 
        $\mathcal{I}$      &
        62.2 & 47.3 & - & 19.8
        \\
        BESTIE$^{\dagger}$~\cite{kim2022beyond}   & HRNet-W48            & $\mathcal{I}$  & 
        61.2 & 51.0 & 31.9 & 26.6
        \\
        
        \midrule
        Ours$^{\dagger}$ & ResNet-50  & 
        $\mathcal{I}$ & \textbf{68.7} & \textbf{55.9} & \textbf{37.1} & \textbf{30.9}
        \\
        \bottomrule
    \end{tabular}
    \caption{Comparison with the state-of-the-art methods on VOC 2012 dataset. $\mathcal{M}$, $\mathcal{S}$, and $\mathcal{I}$ stand for instance-level, instance saliency, and image-level labels, respectively. $\dagger$ indicates training Mask R-CNN for refinement.}
    \label{tab:VOC_dataset}
\end{table*}

\begin{table*}[t!]
    \centering
    \begin{tabular}{c|c|c| c c c c}
        \toprule
        Method & Backbone & Sup. & $AP$ & $mAP_{50}$ & $mAP_{75}$ \\
        \midrule
        \multicolumn{6}{c}{\textbf{\em{COCO val2017}}} \\
        \midrule
        Mask R-CNN~\cite{he2017mask}  & ResNet-101         & $\mathcal{M}$ & 
        35.4 & 57.3 & 37.5 \\
        WS-JDS~\cite{shen2019cyclic}    & VGG-16         & $\mathcal{I}$ & 
        6.1 & 11.7 & 5.5
        \\

        PDSL~\cite{shen2021parallel}     & ResNet18-WS   & $\mathcal{I}$  &
        6.3 & 13.1 & 5.0
        \\    
    
        BESTIE$^{\dagger}$~\cite{kim2022beyond} & HRNet-W48      & $\mathcal{I}$   & 
        14.3 & 28.0 & 13.2
        \\

        \midrule
        Ours & ResNet-50 &
        $\mathcal{I}$ & 11.9 & 22.8 & 11.1
        
        \\
        Ours$^{\dagger}$ & ResNet-50 &
        $\mathcal{I}$ & \textbf{17.0} & \textbf{29.4} & \textbf{17.0}
        \\
        
        \midrule
        \multicolumn{6}{c}{\textbf{\em{COCO test-dev}}} \\
        \midrule
        
        Mask R-CNN~\cite{he2017mask}  & ResNet-101         & $\mathcal{M}$ & 
        35.7 & 58.0 & 37.8 
        \\
        
        Fan et al.$^{\dagger}$~\cite{fan2018associating} & ResNet-101 & $\mathcal{I,S}$ & 
        13.7 & 25.5 & 13.5
        \\
        
        LIID$^{\dagger}$~\cite{liu2020leveraging} & ResNet-50   & $\mathcal{I,S}$ & 
        16.0 & 27.1 & 16.5
        \\
        
        BESTIE$^{\dagger}$~\cite{kim2022beyond}   & HRNet-W48   & $\mathcal{I}$ & 
        14.4 & 28.0 & 13.5
        \\

        \midrule
        Ours & ResNet-50 &
        $\mathcal{I}$ & 12.0 & 23.0 & 11.3
        
        \\
        Ours$^{\dagger}$ & ResNet-50 &
        $\mathcal{I}$ & \textbf{17.2} & \textbf{29.7} & \textbf{17.3}
        \\
        \bottomrule
    \end{tabular}
    \caption{Comparison with the state-of-the-art methods on COCO dataset. $\mathcal{M}$, $\mathcal{S}$, and $\mathcal{I}$ stand for instance-level, instance saliency, and image-level labels, respectively. $\dagger$ indicates training Mask R-CNN for refinement.}
    \label{tab:COCO_dataset}
\end{table*}

\begin{figure*}[t!]
\begin{center}
\includegraphics[width=1\linewidth]{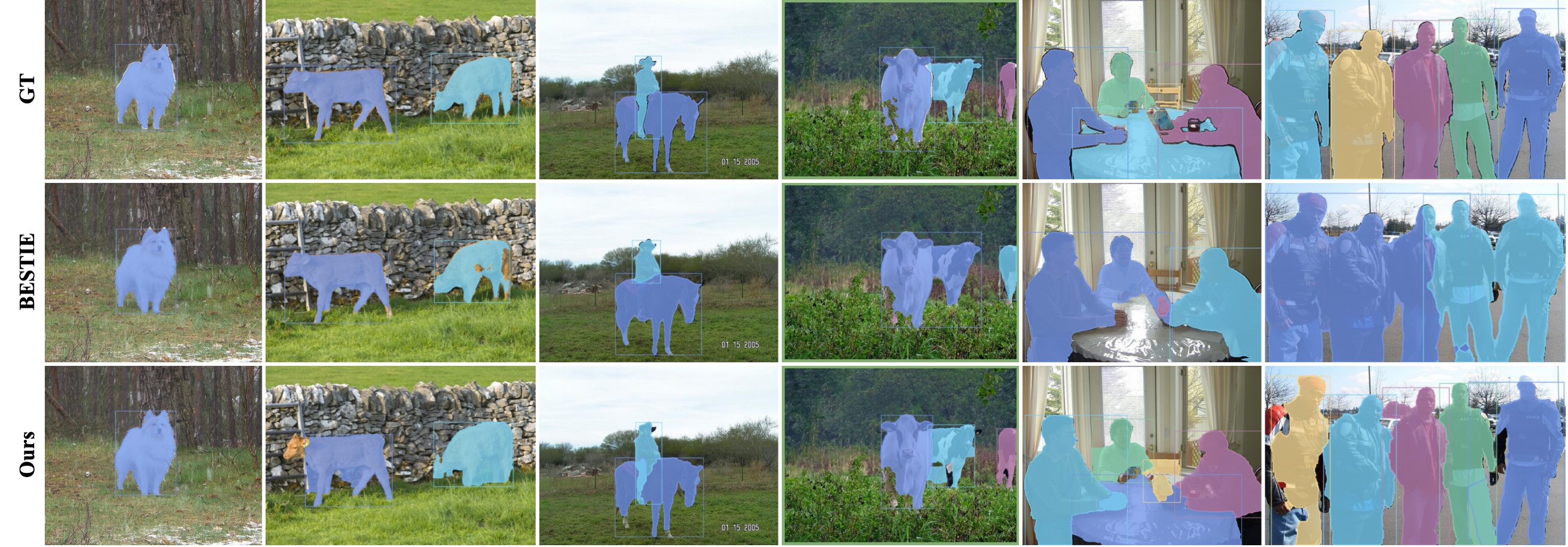}
\end{center}
   \caption{Visualization results on the VOC 2012 dataset. Comparison with BESTIE.}
\label{fig:compare_w_sota}
\end{figure*}

\subsection{Anti-Noise Strategy}
\label{section:Anti-noise strategy}
High-quality pre-computed pseudo labels facilitate the network in achieving faster convergence and better performance. However, noise in these labels can confuse the network and lead to overfitting. Similarly, noisy refined pseudo labels also degrade segmentation performance. To address these issues, we propose an Anti-noise strategy comprising an Anti-noise branch and an Anti-noise sampling strategy.

\textbf{Anti-noise branch}. We adopt the WSDDN branch~\cite{bilen2016weakly} as the Anti-noise branch. Similar to the Refinement branches, it also produces two matrixes $y^0, t^0 \in \mathbb{R}^{N \times \{C+1\}}$ from classification head and Anti-noise head. Then, image-level score can be obtained by $y^I = \sum^{N}_{n=1} (y^0_{n,*} \odot t^0_{n,*})$. We combine Binary Cross-Entropy and PCL~\cite{Tang2020pcl} losses as the objective function.

\begin{align}
    \mathcal{L}_{anti} = &-\frac{1}{C+1} \sum_{c=0}^{C} \{Y_c \log y^I_c + (1 - Y_c) \log (1 - y^I_c)\}  \nonumber \\
             &-\frac{\alpha}{N_{fg} + N_{bg}} \sum_{i=1}^{N^0}\sum_{c=0}^{C} M_i \ s_{i,c} \log \sum_{n \in \mathcal{C}_i} \frac{y^0_{n,c}}{M_i}
\end{align}
Here, the hyper-parameter $\alpha$ is set to $12$ by default. Especially, the Anti-noise head is not directly guided, which enables it to filter out noise in pre-computed pseudo labels.

\textbf{Anti-noise sampling}. Refined pseudo labels also contain noise, as constraints of the CIM strategy are relaxed. Our observations suggest that noisy labels tend to have lower loss weights. Based on this insight, we adopt an Anti-noise sampling strategy that treats loss weights as sampling probabilities to sample pseudo ground truth $P^k$. Specifically, we implement sampling with replacement to filter out noise in refined pseudo labels. 

In summary, the objective function can be written as:
\begin{equation}
    \mathcal{L}_{total}=\mathcal{L}_{anti} + \sum_{k=1}^K \mathcal{L}^k_{ref}
\end{equation}
where $K$ is set to $3$ by default.

\section{Experiments}
\subsection{Datasets and Evaluation Metrics}
Following previous methods, we also evaluate our method on PASCAL VOC 2012~\cite{everingham2010pascal} and MS COCO~\cite{lin2014microsoft} datasets. The VOC 2012 dataset includes 10,582 images for training and 1,449 images for validation, comprising 20 object categories. The COCO dataset comprises 115K training, 5K validation, and 20K testing images across 80 object categories. Following previous methods, we report the mean average precision ($mAP$) with IoU thresholds of 0.25, 0.5, 0.7, and 0.75 for VOC 2012 and report $mAP$ with IoU thresholds from 0.5 to 0.95 for COCO.


\subsection{Implementation Details}
Our method is implemented in PyTorch and experiments are conducted on an Nvidia RTX 3090. We use COB~\cite{maninis2017convolutional} method to generate proposals for all experiments and utilize ResNet50~\cite{he2016resnet} as the backbone. As we using $mAP_{25}$ and $mAP_{50}$ as evaluation metrics, we set classification $\tau_{cls}$ and integrity $\tau_{iou}$ thresholds to $0.25$ and $0.5$, respectively. The cascaded threshold $\tau_{cas}$ is set to $0.1$. $\tau_{nms}$, and $p_{seed}$ in Algorithm~\ref{alg:CIM} are set to $\tau_{cls}$, and $0.1$, respectively. Containment threshold $\tau_{con}$ is set to $0.85$ following SoS~\cite{Sui2021SoS}. Although there are many hyper-parameters, we only tune values of $\tau_{cas}$ and $p_{seed}$.

During training, we use the SGD optimization algorithm with an initial learning rate of $2.5 \times {10}^{-4}$ and a weight decay of $5 \times {10}^{-4}$. We adopt a step learning rate decay schema with a decay weight of $0.1$ and set the mini-batch size to $4$. The total number of training iterations is $4.5 \times {10}^{4}$ for the VOC 2012 dataset and $24 \times {10}^{4}$ iterations for the COCO dataset. For data augmentation, we apply five image scales $\{480, 576, 688, 864, 1200\}$ with random horizontal flips for both training and testing. During testing, we employ the product of classification and integrity scores as the output of each Refinement branch and average these outputs as the final scores. Following previous methods, we also generate pseudo labels from our method for training Mask R-CNN.

\subsection{Comparison With State-of-the-Art}
We compare the performance of our method with previous state-of-the-art WSIS methods, as shown in Table~\ref{tab:VOC_dataset} and Table~\ref{tab:COCO_dataset}. For a fair comparison, we also report the results obtained with different backbones, i.e., VGG-16~\cite{Simonyan2015vgg} and HRNet-W48~\cite{Wang2021HRNet}. Note that MIST is originally an object detection method, and we adapt it to the WSIS task for comparison.

Our proposed method outperforms all previous methods on both datasets, achieving higher performance than LIID~\cite{liu2020leveraging} even without the use of instance saliency labels, demonstrating that such labels are not necessary for further advancement in WSIS. Additionally, our method demonstrates a $4.9 \%$ and $1.7 \%$ improvement in terms of $mAP_{50}$ on the VOC 2012 and COCO, respectively, when compared with BESTIE~\cite{kim2022beyond}. Thanks to impressive semantic segmentation maps produced by WSSS, BESTIE achieves excellent performance on simple images, i.e., single instance, multiple non-adjacent instances, and multiple instances of different categories. However, BESTIE tends to produce grouping instances results caused by its strengthening constraints, illustrated in the second row of Figure~\ref{fig:compare_w_sota}. In contrast, we follow the proposal-based paradigm which always results in redundant segmentation rather than grouping instances. Hence, we propose CIM to mine complete instances, illustrated in the third row of Figure~\ref{fig:compare_w_sota}. 

Although my approach may appear overly complex, it is implemented using only one linear layer per head, and its training can be completed in a few hours, as shown in Table~\ref{tab:Time spent}. Meanwhile, the code implementation of CIM is simple.

\subsection{Ablation Study}
We conduct several ablation studies on the VOC 2012 dataset to evaluate the effectiveness of each component. In these studies, we use ResNet-50 as the backbone and do not apply Mask R-CNN to save time. When the CIM strategy is unavailable, we adopt MIST~\cite{ren-cvpr2020} strategy to replace it and adapt $\tau_{cls}$ and $\tau_{nms}$ to $0.5$. 

\begin{table}[t!]
    \centering
    \begin{tabular}{c c c}
        \toprule
        Dataset & Backbone & Time $(h)$ \\
        \midrule
        VOC 2012 & VGG-16 & 4.7
        \\
        VOC 2012 & ResNet-50 & 6.5
        \\ 
        VOC 2012 & HRNet-W48 & 30.8
        \\ 
        COCO & ResNet-50 &  37.6
        \\
        \bottomrule
    \end{tabular}
    \caption{Time spent on different configurations.}
    \label{tab:Time spent}
\end{table}

\begin{table}[t!]
    \centering
    \begin{tabular}{c c c c | c c}
        \toprule
        AGPL & MaskIoU & CIM & Cas & $mAP_{50}$ & $mAP_{75}$\\
        \midrule
        
        & & & & 38.1 & 17.5
        \\
        
        \ding{52} & & & & 49.2 & 20.6
        \\ 

        \ding{52} & \ding{52} & & & 48.8 & 21.4 
        \\ 

        \ding{52} & \ding{52} & \ding{52} & & 50.1 & 23.8
        \\ 

        \ding{52} & \ding{52} & \ding{52} & \ding{52} & 51.1 & 26.1
        \\
        \bottomrule
    \end{tabular}
    \caption{Impact of each component: AGPL, MaskIoU head, CIM strategy, and cascaded threshold.}
    \label{tab:ablation_study_each_componet}
\end{table}

\begin{figure}[t!]
    \begin{center}
    \includegraphics[width=0.95\linewidth]{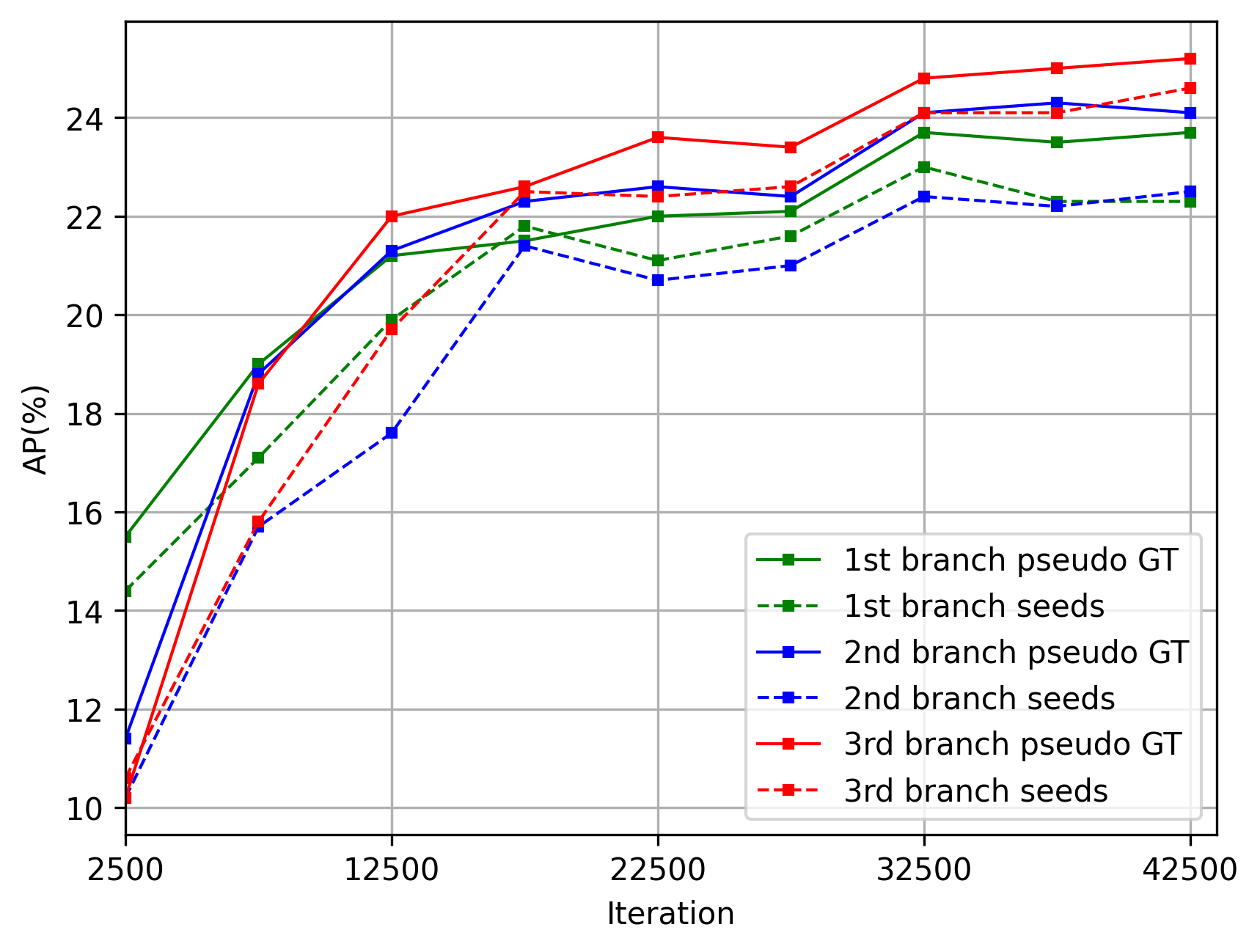}
    \end{center}
    \caption{Performance of pseudo GT and seeds generated by CIM.}
    \label{fig:CIM_statistics}
\end{figure}

\subsubsection{Impact of AGPL}
As shown in Table~\ref{tab:ablation_study_each_componet}, the use of AGPL results in an improvement in segmentation performance, with an increase of $49.2\%$ and $20.6\%$ in terms of $mAP_{50}$ and $mAP_{75}$, respectively. Although AGPL produces reliable pre-computed pseudo labels, it only provides coarse complete instance cues, leading to a significant improvement in $mAP_{50}$ but little gain in $mAP_{75}$.

\subsubsection{Impact of MaskIoU and CIM}
As shown in Table~\ref{tab:ablation_study_each_componet}, the performance is slightly improved when the MaskIoU heads are applied alone because the distribution of refined pseudo labels does not change. By employing the CIM strategy, our method achieves a $1.3\%$ and $2.4\%$ improvement in terms of $mAP_{50}$ and $mAP_{75}$, respectively. This demonstrates that our method can effectively mine complete instances, resulting in improved performance for stricter metrics, i.e., $mAP_{75}$.

\subsubsection{Impact of Cascaded Threshold}
Furthermore, the cascaded threshold design allows our method to operate in a coarse-to-fine manner. As shown in the last row of Table~\ref{tab:ablation_study_each_componet}, the cascaded threshold results in a $1.0\%$ and $2.3\%$ improvement in terms of $mAP_{50}$ and $mAP_{75}$, respectively, as it guides deeper Refinement branches to focus on more complete instances.

We also evaluate the seeds and pseudo ground truth generated by CIM, as shown in Figure~\ref{fig:CIM_statistics}. It is obvious that pseudo ground truth outperforms seeds and deeper Refinement branch performs better. As reported in the Table~\ref{tab:ablation_study_cascade threshold}, the effect of different cascade thresholds is small.

\begin{table}[t!]
    \centering
    \begin{tabular}{c|c c c c c}
        \toprule
        $\tau_{cas}$ & $0.00$ & $0.05$ & $0.10$ & $0.15$ & $0.20$ \\
        \midrule
        $mAP_{50}$ & 50.1 & 51.6 & 51.1 & 50.4 & 50.5
        \\ 
        $mAP_{75}$ & 23.8 & 25.9 & 26.1 & 26.8 & 26.4
        \\
        \bottomrule
    \end{tabular}
    \caption{Impact of the cascaded threshold value.}
    \label{tab:ablation_study_cascade threshold}
\end{table}

\begin{table}[t!]
    \centering
    \begin{tabular}{c|c c c c}
        \toprule
        $p_{seed}$ & $0.05$ & $0.10$ & $0.15$ & $0.20$ \\
        \midrule
        \ding{56} & 50.5 & 49.4 & 48.6 & 47.8
        \\ 

        \ding{52} & 51.5 & 51.1 & 50.9 & 51.1
        \\ 
 
        \bottomrule
    \end{tabular}
    \caption{Impact of Anti-noise sampling. \ding{56} and \ding{52} mean without and with sampling, respectively.}
    \label{tab:ablation_study_sample}
\end{table}

\subsubsection{Impact of Anti-Noise Strategy}
Following UFO$^2$\cite{ren2020ufo}, we treat the pre-computed pseudo labels as ground truth and apply a KL-divergence loss on the Anti-noise head, which reduces its performance to $50.4\%$ in terms of $mAP_{50}$. This result suggests that redundant parameters in the Anti-noise head effectively filter out noise in pre-computed pseudo labels. 

To further evaluate the impact of Anti-noise sampling, we introduce noise by increasing the number of seeds in the CIM process. As shown in Table~\ref{tab:ablation_study_sample}, the use of Anti-noise sampling increases the robustness of our method.

\section{Conclusion}

In this paper, we propose a proposal-based approach in an online refinement manner to address the redundant segmentation problem. Our method incorporates the MaskIoU head and utilizes the CIM strategy to mine complete instances without resorting to RW or CRF. Additionally, we propose the Anti-noise strategy to filter out noise in pseudo labels. Our approach demonstrates state-of-the-art performance on the VOC 2012 and COCO datasets. 
Moving forward, we expect to investigate methods to enhance the robustness of the model without resorting to Anti-noise sampling strategy that may complicate the analysis.

\section*{Acknowledgments}
This work was funded by Natural Science Foundation of China under Grants No. 62076099 and No. 61703166.

\bibliographystyle{named}
\bibliography{ijcai23}

\end{document}